\documentclass[letterpaper]{article}

\newcommand{\modelname}{\textsc{GoalCover}}

\newcommand{\phasetitle}{\sffamily\footnotesize\bfseries}

\ifdefined\aaaianonymous
    \usepackage[submission]{aaai2026}
\else
    \usepackage{aaai2026}
\fi

\usepackage{times}
\usepackage{helvet}
\usepackage{courier}
\usepackage[hyphens]{url}
\usepackage{graphicx}
\usepackage{natbib}
\usepackage{caption}
\usepackage{amsmath}
\usepackage{amsfonts}
\usepackage{amssymb}
\usepackage{amsthm}
\usepackage{enumitem}
\usepackage{mdframed}
\usepackage{booktabs}
\usepackage{array}
\usepackage{longtable}
\usepackage{xcolor}
\usepackage{algorithm}
\usepackage{algorithmic}
\usepackage{multirow}

\usepackage{tikz}
\usepackage{pifont}
\usetikzlibrary{positioning,shapes.geometric,arrows.meta,calc,fit,backgrounds}

\definecolor{gcblue}{RGB}{46,92,138}
\definecolor{gcblue_l}{RGB}{225,236,247}
\definecolor{gcteal}{RGB}{91,154,160}
\definecolor{gcteal_l}{RGB}{222,240,242}
\definecolor{gcgreen}{RGB}{74,124,89}
\definecolor{gcgreen_l}{RGB}{222,239,228}
\definecolor{gcred}{RGB}{181,73,74}
\definecolor{gcred_l}{RGB}{248,225,225}
\definecolor{gcorange}{RGB}{200,130,50}
\definecolor{gcorange_l}{RGB}{250,235,213}
\definecolor{gcpurple}{RGB}{120,80,140}
\definecolor{gcpurple_l}{RGB}{235,225,242}
\definecolor{gcgray}{RGB}{120,120,120}
\definecolor{gctxt}{RGB}{40,40,40}

\newtheorem{definition}{Definition}

\ifdefined\aaaianonymous
    \title{Diagnosing Capability Gaps in Fine-Tuning Data
}
\else
    \title{Diagnosing Capability Gaps in Fine-Tuning Data
}
\fi

\ifdefined\aaaianonymous
\author{
    Anonymous Submission
}
\affiliations{
    
}
\else
\author{
    Saeid Asgari Taghanaki,
    Rakshanda Agarwal\equalcontrib,
    Bruce Sun\equalcontrib,
    Rohan Jha,
    Elias Stengel-Eskin,\\
    Sara Malvar,
    Rui Ying,
    Yifei Xu,
    Guilherme Potje, 
    Tusher Chakraborty,
    Leonardo de Oliveira Nunes,
    Ranveer Chandra,
    Emre Kiciman
}
\affiliations{
    Microsoft
}
\fi

\begin{document}

\maketitle

\begin{abstract}
Fine-tuning large language models (LLMs) for domain-specific tasks requires training datasets that comprehensively cover the target capabilities a practitioner needs. Yet identifying \emph{which} capabilities a dataset fails to support, and doing so \emph{before} an expensive fine-tuning run, remains a largely unsolved problem. We introduce \modelname{}, a framework that helps practitioners systematically detect capability gaps in fine-tuning datasets through interactive goal decomposition and automated coverage assessment. \modelname{} guides a practitioner through structured decomposition of a high-level goal into atomic, independently evaluable subgoals; assigns each training sample an LLM-based alignment score against every subgoal; and surfaces missing capabilities through automated analysis of low-scoring sample explanations. We validate the framework along two complementary axes. First, through controlled corruption experiments across three domains (medical QA, legal summarization, code generation), we show that \modelname{} reliably distinguishes targeted from non-targeted capability impacts: target subgoals degrade by 25.6\% on average versus 2.1\% for non-target subgoals (Cohen's $d{=}1.24$). Second, we demonstrate downstream utility on a financial-summarization Reinforcement Fine-Tuning (RFT) task with Qwen-3-14B: training on \modelname{}-filtered data improves the LLM-judge reward from 3.77 to 4.12 (out of 5) over the unfiltered baseline, and combining filtered data with goal-conditioned synthetic samples yields the strongest result (4.20). The two results together show that \modelname{} works as a practical pre-fine-tuning diagnostic: it detects capability gaps and produces concrete signal for closing them.
\end{abstract}

\ifdefined\aaaianonymous
\else
\begin{links}
\end{links}
\fi

\section{Introduction}

\begin{figure*}[t]
\centering
\begin{tikzpicture}[
    every node/.style={font=\sffamily},
    goalbox/.style={
        rectangle, rounded corners=4pt,
        draw=gcblue, line width=0.7pt,
        fill=gcblue_l,
        inner xsep=10pt, inner ysep=7pt,
        text width=7.6cm, align=center,
        font=\sffamily\small, text=gctxt
    },
    subgoalcard/.style={
        rectangle, rounded corners=3pt,
        draw=gcgray, line width=0.5pt,
        fill=white,
        inner xsep=4pt, inner ysep=5pt,
        text width=2.2cm, align=center,
        font=\sffamily\scriptsize, text=gctxt,
        minimum height=0.95cm
    },
    outcome_ok/.style={
        rectangle, rounded corners=2.5pt,
        fill=gcgreen_l, draw=gcgreen, line width=0.5pt,
        inner xsep=5pt, inner ysep=3pt,
        font=\sffamily\scriptsize\bfseries,
        text=gcgreen, minimum width=2.2cm, align=center
    },
    outcome_fail/.style={
        rectangle, rounded corners=2.5pt,
        fill=gcred_l, draw=gcred, line width=0.5pt,
        inner xsep=5pt, inner ysep=3pt,
        font=\sffamily\scriptsize\bfseries,
        text=gcred, minimum width=2.2cm, align=center
    },
    arr/.style={->, >={Stealth[length=4pt,width=4pt]}, draw=gcgray, line width=0.5pt},
    annot/.style={font=\sffamily\scriptsize\itshape, text=gcgray}
]
\node[goalbox] (goal) at (0, 4.6) {%
    \textbf{Practitioner goal} \\[2pt]
    \emph{``Build a medical QA system for clinical decision support.''}%
};
\node[annot, anchor=east] at (-3.6, 3.6) {decompose into atomic subgoals};
\draw[arr] (goal.south) -- ++(0,-0.7);

\def\rowy{2.65}
\node[subgoalcard] (s1) at (-5.6, \rowy) {Clinical\\reasoning};
\node[subgoalcard] (s2) at (-2.8, \rowy) {Cardiology\\expertise};
\node[subgoalcard] (s3) at (0,    \rowy) {Drug\\information};
\node[subgoalcard] (s4) at (2.8,  \rowy) {Safety\\warnings};
\node[subgoalcard] (s5) at (5.6,  \rowy) {Evidence\\citations};
\foreach \tgt in {s1,s2,s3,s4,s5} {
    \draw[arr] (goal.south) ++(0,-0.45) -- (\tgt.north);
}

\def\bary{1.55}
\def\barh{0.18}
\def\barw{2.2}
\node[annot, anchor=east] at (-7, \bary+\barh/2) {Coverage:};
\newcommand{\covbar}[2]{%
    \draw[fill=gray!12, draw=gcgray!40, line width=0.3pt]
        (#1-\barw/2, \bary) rectangle (#1+\barw/2, \bary+\barh);
    \pgfmathsetmacro{\fillw}{#2*\barw}
    \pgfmathparse{ifthenelse(#2<0.4,1,0)}
    \ifnum\pgfmathresult=1
        \draw[fill=gcred!75, draw=none]
            (#1-\barw/2, \bary) rectangle (#1-\barw/2+\fillw, \bary+\barh);
    \else
        \draw[fill=gcgreen!75, draw=none]
            (#1-\barw/2, \bary) rectangle (#1-\barw/2+\fillw, \bary+\barh);
    \fi
    \node[font=\sffamily\tiny, text=gctxt, anchor=west]
        at (#1+\barw/2+0.05, \bary+\barh/2) {#2};
}
\covbar{-5.9}{0.85}
\covbar{-3.}{0.18}
\covbar{-.1}{0.12}
\covbar{2.8}{0.62}
\covbar{5.7}{0.71}

\def\outy{0.55}
\node[annot, anchor=east] at (-6.7, \outy) {After fine-tuning:};
\node[outcome_ok]   (o1) at (-5.6, \outy) {\ding{51}\ Reliable};
\node[outcome_fail] (o2) at (-2.8, \outy) {\ding{55}\ FAILS};
\node[outcome_fail] (o3) at (0,    \outy) {\ding{55}\ FAILS};
\node[outcome_ok]   (o4) at (2.8,  \outy) {\ding{51}\ Reliable};
\node[outcome_ok]   (o5) at (5.6,  \outy) {\ding{51}\ Reliable};
\foreach \xc in {-5.6,-2.8,0,2.8,5.6} {
    \draw[dotted, draw=gcgray!50, line width=0.4pt]
        (\xc, \bary) -- (\xc, \outy+0.22);
}
\node[font=\sffamily\footnotesize, anchor=center, text=gctxt, text width=15.5cm, align=center]
    at (0, -0.55) {%
    \textcolor{gcgray}{Average coverage score $\approx 0.50$, \emph{looks reasonable.}}\;%
    \textcolor{gcred}{Subgoal-level coverage reveals gaps that surface only as predictable production failures.}%
};
\end{tikzpicture}
\caption{Capability gaps in fine-tuning datasets. The average coverage score across subgoals can look adequate while individual subgoals remain severely undersupported, surfacing only as failures in production. \modelname{} provides the missing subgoal-level visibility before fine-tuning begins.}
\label{fig:motivation}
\end{figure*}
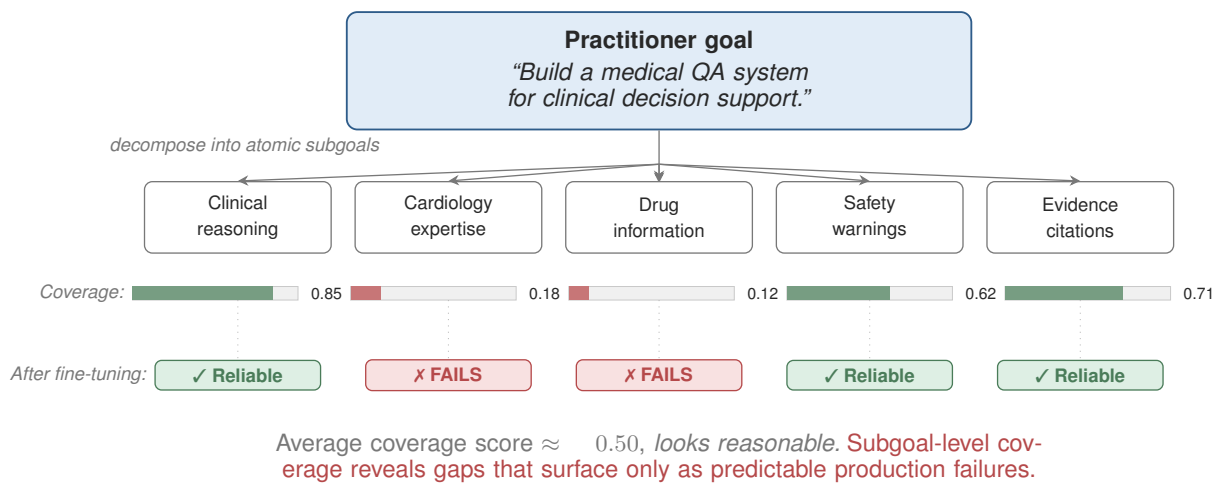

The success of domain-specific LLM fine-tuning depends critically on whether the training dataset actually covers the target capabilities the practitioner cares about \citep{wei2022finetuned, ouyang2022training}. As LLMs are deployed in specialized settings such as clinical decision support, legal document processing, code generation, and scientific research, a recurring problem emerges: a practitioner can articulate a high-level goal (e.g., ``generate accurate medical answers with appropriate safety considerations'' or ``summarize legislation while preserving monetary and regulatory detail''), but has no principled way to determine \emph{whether their dataset can support that goal} until after a costly fine-tuning run reveals the gap. Figure~\ref{fig:motivation} illustrates the failure mode: aggregate dataset alignment can look reasonable while individual subgoals remain undersupported, surfacing only as predictable production failures.

Existing dataset-quality methods leave this gap mostly unaddressed. Statistical metrics such as perplexity, BLEU, and distributional analysis provide aggregate signal but cannot localize which target capability is undersupported \citep{swayamdipta2020dataset, pleiss2020identifying, marion2023less}. Expert review is valuable but expensive, subjective, and difficult to scale across the multi-dimensional objectives of modern applications. Post-hoc performance analysis only reveals problems after fine-tuning is complete, forcing iterative cycles of training, evaluation, and dataset refinement that consume substantial compute.

This evaluation gap becomes consequential as practitioners invest in fine-tuning for high-stakes applications. A medical assistant that performs well on general clinical questions but fails on cardiology cases due to insufficient cardiac training data could mislead clinicians. A legal summarizer that captures most of a bill but consistently omits monetary appropriations could produce summaries that are legally insufficient. These represent systematic \emph{capability gaps} rather than mere performance issues: specific sub-components of a goal that lack sufficient training support to enable reliable model behavior.

The challenge is compounded by the structure of modern fine-tuning goals. Unlike traditional ML tasks with a single defined metric, practical fine-tuning objectives combine domain knowledge, safety, formatting, and behavioral expectations. A practitioner fine-tuning a medical QA model needs accurate answers \emph{and} appropriate clinical terminology \emph{and} safety warnings for contraindications \emph{and} evidence-based reasoning across multiple specialties. Assessing dataset coverage across these heterogeneous requirements demands a method that operates at subgoal granularity.

Recent advances in LLM-based evaluation provide useful foundations \citep{zheng2023judging, liu2023geval, li2024generative, kim2024prometheus2}. Inspired by the success of LLM-as-a-Judge across diverse tasks, we are, to our knowledge, the first to apply this paradigm to dataset-goal coverage assessment, with explicit attention to goal specification, scoring criteria, and validation methodology.

\begin{figure*}[t]
\centering
\begin{tikzpicture}[
    every node/.style={font=\sffamily},
    phasebox/.style={
        rectangle, rounded corners=3pt,
        draw=#1, line width=0.6pt, fill=white,
        inner xsep=6pt, inner ysep=8pt,
        text width=3.4cm, align=center,
        font=\sffamily\scriptsize, text=gctxt,
        minimum height=2.7cm
    },
    phasebox/.default=gcblue,
    phasetitle/.style={font=\sffamily\footnotesize\bfseries},
    flowarrow/.style={->, >={Stealth[length=5.5pt,width=5pt]}, draw=gcgray, line width=0.7pt},
    plainline/.style={draw=gcgray, line width=0.7pt},
    iolabel/.style={font=\sffamily\scriptsize\itshape, text=gctxt!85},
    valbox/.style={
        rectangle, rounded corners=4pt,
        draw=#1, line width=0.7pt,
        inner xsep=8pt, inner ysep=7pt,
        text width=5.4cm, align=center,
        font=\sffamily\footnotesize, text=gctxt
    },
    valbox/.default=gcgreen
]
\def\phaseY{2.0}
\def\spacing{4.0}
\node[
    rectangle, rounded corners=3pt,
    draw=gcpurple, line width=0.5pt, fill=gcpurple_l,
    inner xsep=6pt, inner ysep=4pt,
    text width=2.4cm, align=center,
    font=\sffamily\scriptsize, text=gctxt
] (input) at (-3.1*\spacing/2, \phaseY+2.55) {%
    \textbf{Practitioner}\quad\textit{goal} $\mathcal{G}$
};
 \node[phasebox=black] (p1) at (-3.1*\spacing/2, \phaseY) {%
    {\phasetitle{Phase 1}}\\
    {\phasetitle{Goal Clarification}}\\[3pt]
    \textcolor{gcgray}{\rule{1.6cm}{0.3pt}}\\[3pt]
    Iterative Q$\,\to\,$R loop\\
    until atomic\\decomposition\\[4pt]
    {\scriptsize $\mathcal{S}=f_{\mathrm{decompose}}(\mathcal{G},Q,R)$}
};
\node[phasebox=black] (p2) at (-\spacing/2, \phaseY) {%
    {\phasetitle{Phase 2}}\\
    {\phasetitle{Coverage Assessment}}\\[3pt]
    \textcolor{gcgray}{\rule{1.6cm}{0.3pt}}\\[3pt]
    Score every\\$(x_i,y_i,s)$ pair:\\[2pt]
    {\scriptsize $A(x_i,y_i,s){\in}[0,1]$}\\
    {\scriptsize $+$ explanation $r_i$}
}; 
\node[phasebox=black] (p3) at (1.2*\spacing/2, \phaseY) {%
    {\phasetitle{Phase 3}}\\
    {\phasetitle{Gap Analysis}}\\[3pt]
    \textcolor{gcgray}{\rule{1.6cm}{0.3pt}}\\[3pt]
    For $s$ with $\bar A(s){<}\tau$:\\
    aggregate explanations\\
    of low-scoring samples\\[3pt]
    {\scriptsize $\mathrm{Gap}_s=f_{\mathrm{analyze}}(\mathcal{R}_{\mathrm{low}})$}
};
\node[phasebox=black] (p4) at (3.5*\spacing/2, \phaseY) {%
    {\phasetitle{Phase 4}}\\
    {\phasetitle{Recommendation}}\\[3pt]
    \textcolor{gcgray}{\rule{1.6cm}{0.3pt}}\\[3pt]
    Issues + fixes per gap $s$,\\plus goal-conditioned\\synthetic data\\[3pt]
    {\scriptsize $\mathrm{Rec}_s=f_{\mathrm{recommend}}(\mathrm{Gap}_s)$}
};
\draw[flowarrow] (input.south) -- (p1.north);
\draw[flowarrow] (p1.east) -- node[iolabel, above, yshift=1pt]{$\mathcal{S}$} (p2.west);
\draw[flowarrow] (p2.east) -- node[iolabel, above, yshift=1pt]{$A,\, r$} (p3.west);
\draw[flowarrow] (p3.east) -- node[iolabel, above, yshift=1pt]{$\mathrm{Gap}_s$} (p4.west);
\begin{scope}[on background layer]
    \node[
        draw=gcgray!30, line width=0.4pt, dashed,
        rounded corners=4pt,
        fit=(p1)(p2)(p3)(p4),
        inner sep=10pt
    ] (pipelinegroup) {};
\end{scope}
\node[
    fill=white, font=\sffamily\footnotesize\itshape, text=gcgray,
    inner xsep=4pt, inner ysep=1pt
] (grouplabel) at (pipelinegroup.north) {\textsc{GoalCover} pipeline};
\node[font=\sffamily\scriptsize\itshape, text=gcgray]
    (outlbl) at (0, \phaseY - 1.8) {pre-fine-tuning diagnostic};
\def\valY{-2.0}
\node[valbox=gcgreen, fill=gcgreen_l] (val1) at (-3.2, \valY) {%
    \textbf{\textcolor{gcgreen}{Validation 1: Detection}} \\[1pt]
    \scriptsize Controlled corruption (\S5)\\
    \scriptsize Target vs.\ non-target degradation: \\
    \scriptsize \textbf{25.6\%} vs.\ \textbf{2.1\%},\quad Cohen's $d{=}1.24$
};
\node[valbox=gcgreen, fill=gcgreen_l] (val2) at (4, \valY) {%
    \textbf{\textcolor{gcgreen}{Validation 2: Downstream Utility}}\\[1pt]
    \scriptsize RFT, Qwen-3-14B + GRPO (\S7)\\
    \scriptsize Filtered: \textbf{3.77\,$\to$\,4.12}\\
    \scriptsize Filtered $+$ goal-cond.\ synth.: \textbf{4.20}
};
\coordinate (trunkL) at (-3, \valY + 1.05);
\coordinate (trunkC) at (0.4,    \valY + 1.05);
\coordinate (trunkR) at (3.8,  \valY + 1.05);
\draw[plainline] (pipelinegroup.south) -- (trunkC);
\draw[plainline] (trunkL) -- (trunkR);
\draw[flowarrow] (trunkL) -- (val1.north);
\draw[flowarrow] (trunkR) -- (val2.north);
\end{tikzpicture}
\caption{Overview of \modelname{}. The four-phase pipeline takes a practitioner's high-level goal $\mathcal{G}$, decomposes it into atomic subgoals via an interactive clarification loop, scores every (sample, subgoal) pair, aggregates evaluator explanations from low-scoring samples into capability gaps, and produces remediation recommendations including goal-conditioned synthetic data. We validate the pipeline along two complementary axes: a controlled corruption study (Section~\ref{sec:val}) establishes detection validity, and a downstream RFT study (Section~\ref{sec:rft}) establishes utility.}
\label{fig:method}
\end{figure*}
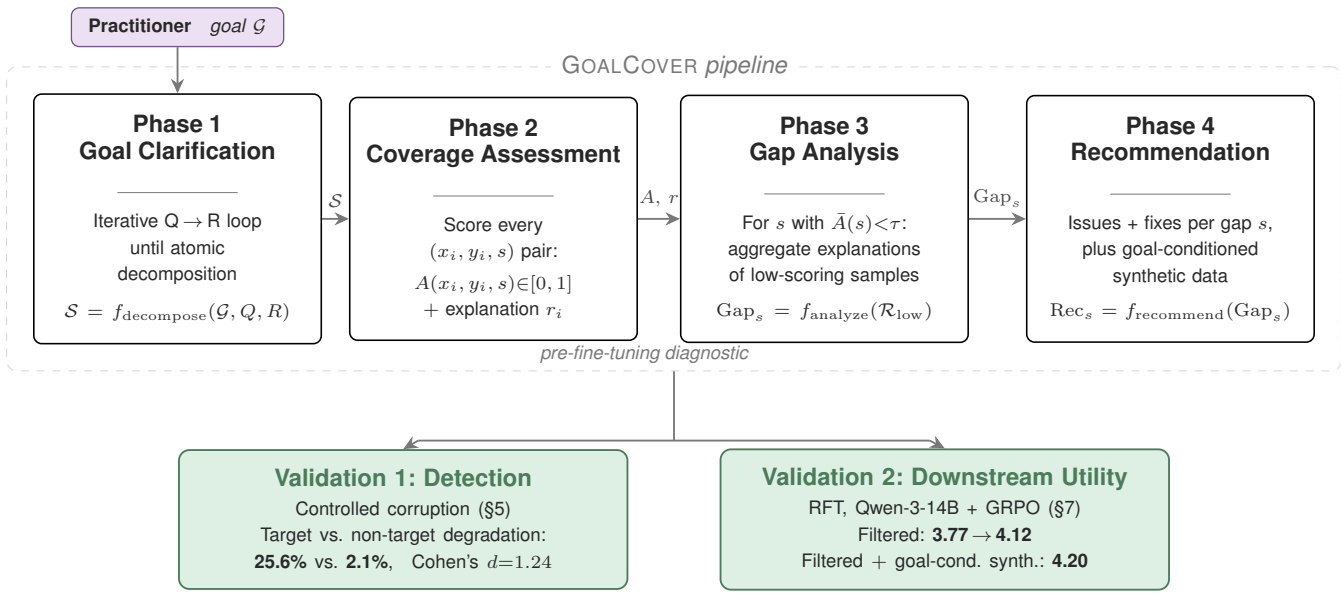

\modelname{} (Figure~\ref{fig:method}) operates in two phases: (1) an interactive goal-clarification system that helps practitioners decompose objectives into specific, testable subgoals, and (2) an automated coverage pipeline that scores each training sample against each subgoal and surfaces missing capabilities through analysis of the evaluator's structured explanations. To validate the framework's detection mechanism without requiring full user studies, we design controlled corruption experiments that simulate realistic gap scenarios by selectively removing target content; we then independently validate downstream utility on a Reinforcement Fine-Tuning (RFT) task where data filtered by \modelname{} measurably improves the trained model. \modelname{}-filtered data raises the RFT reward from 3.77 to 4.12 (Section~\ref{sec:rft}), with the strongest configuration (filtered plus goal-conditioned synthetic data) reaching 4.20.

\modelname{} addresses three research questions: (i) How can complex fine-tuning goals be decomposed into evaluable subcomponents through structured practitioner interaction? (ii) Can LLM-based evaluators reliably assess training-sample alignment with specific capability requirements? (iii) Does goal-conditioned coverage assessment yield useful signal for improving fine-tuning outcomes? We address these questions through the following contributions, which together define \modelname{} as a structured framework for decomposing fine-tuning goals, evaluating data--capability alignment, and validating the resulting gap signal.

\begin{enumerate}
\item \textbf{Structured goal--subgoal decomposition.} A principled methodology for breaking complex goals into atomic, independently evaluable subgoals that map to specific model capabilities. The decomposition is structured in the sense that it (a) is produced through a fixed elicitation protocol (clarifying-question-answering loop) rather than an ad-hoc prompt, (b) enforces atomicity and disjointness constraints (Eqs.~\ref{eq:goal_decomp}--\ref{eq:atomic_subgoals}), and (c) is itself indirectly validated by the controlled corruption experiments: a poor decomposition would not yield the clean target/non-target separation we observe.

\item \textbf{LLM-based alignment assessment.} A scoring framework using anchored-rubric prompts that produces both a quantitative alignment score and a structured explanation of why a sample does or does not support a given subgoal. The explanations are reused as input to the gap-analysis stage.

\item \textbf{Adversarial validation methodology.} A controlled corruption protocol (in spirit, an ablation study on data) that selectively removes content known to support specific subgoals and verifies that the detection mechanism flags genuine capability dependencies rather than diffuse degradation.

\item \textbf{Cross-domain empirical validation.} Evaluation across medical QA, legal summarization, and code generation, demonstrating reliable target-versus-non-target separation (mean 25.6\% vs.\ 2.1\% degradation; Cohen's $d{=}1.24$).

\item \textbf{Downstream RFT utility.} On a financial-summarization RFT task with Qwen-3-14B trained via GRPO, \modelname{}-filtered data improves the LLM-judge reward from 3.77 to 4.12 (out of 5), with subgoal-level analysis showing consistent gains across all coverage subgoals.

\item \textbf{Goal-conditioned synthetic data generation.} A targeted synthesis strategy that uses subgoal-level gap signal and evaluator explanations to generate training samples explicitly addressing detected deficiencies. Combining filtered real data with goal-conditioned synthetic data achieves the strongest RFT result (4.20).

\item \textbf{Practical curation pipeline.} An end-to-end workflow giving practitioners specific, actionable feedback about dataset adequacy and targeted recommendations for closing identified gaps.
\end{enumerate}

\section{Related Work}

\paragraph{Dataset quality assessment and curation.} Traditional approaches to training-data evaluation have focused on statistical properties, distributional analysis, and label quality. \citet{swayamdipta2020dataset} introduced data maps to characterize ambiguous, easy, and hard training examples; \citet{pleiss2020identifying} examined how training dynamics expose data-quality issues; \citet{northcutt2021confident} developed confident learning for systematic label-error detection. More recent work has explored automated assessment: \citet{koh2021wilds} introduced distribution-shift benchmarks, and the data-centric AI literature provides comprehensive surveys \citep{zha2023datacentric}. Recent work has continued to use perplexity- and distribution-based selection signals at scale \citep{marion2023less}, motivating the question of how to complement them with goal-aware diagnostics. These methods primarily address distributional and labeling issues rather than goal-specific capability gaps.

\paragraph{Fine-tuning data selection.} Selecting high-quality training data for fine-tuning has received considerable attention. \citet{xia2024less} demonstrated that careful selection achieves strong performance with substantially fewer examples; \citet{li2024quantity} studied quantity--quality tradeoffs in instruction tuning; \citet{zhou2023lima} showed that small, carefully curated datasets can match much larger collections; \citet{tirumala2022memorization} investigated memorization patterns to guide selection. \modelname{} complements these by providing capability-level gap signal that can inform selection \emph{before} fine-tuning begins.

\paragraph{LLM-based evaluation.} The use of LLMs as evaluators has emerged as a powerful paradigm \citep{zheng2023judging, liu2023geval}. \citet{kim2024prometheus2} developed evaluation protocols with attention to prompt design and bias mitigation; \citet{dubois2024length} identified and addressed systematic biases such as length bias; \citet{fu2023gptscore} explored reference-free evaluation. \citet{li2026grading} systematically studied how the choice of grading scale (e.g., 0--5 vs.\ 1--10) affects human--LLM agreement, finding that scale design materially shifts judge calibration. Recent work has examined confidence estimation \citep{lin2022teaching} and meta-evaluation. We build on these foundations by applying LLM evaluation specifically to dataset--goal alignment, with attention to anchored-rubric scoring and corruption-based validation.

\paragraph{Interactive and human--AI collaborative evaluation.} \citet{stiennon2020learning} demonstrated human-feedback integration for training-data improvement; \citet{ouyang2022training} scaled this for instruction following; \citet{bai2022constitutional} introduced constitutional approaches combining human feedback with AI-generated evaluation; \citet{wang2023selfinstruct} explored self-supervised instruction-data generation; \citet{xu2023wizardlm} developed iterative refinement using LLM feedback. Where these approaches focus on \emph{post-hoc} improvement through feedback, \modelname{} provides systematic \emph{pre-training} gap detection through interactive goal clarification.

\paragraph{Position relative to prior work.} \modelname{} departs from existing methods by focusing on proactive, capability-specific assessment rather than reactive performance analysis or general data-quality metrics. Unlike post-hoc evaluation that reveals problems after training, \modelname{} enables practitioners to identify and address capability gaps before committing to a fine-tuning run, with the framework's design (Section~\ref{sec:method}) and its validation (Sections~\ref{sec:val}--\ref{sec:rft}) addressed as separate contributions.

\section{Method}\label{sec:method}

The \modelname{} pipeline (Figure~\ref{fig:method}) consists of four phases, applied in sequence: \emph{Goal Clarification}, \emph{Coverage Assessment}, \emph{Gap Analysis}, and \emph{Recommendation}. We first formalize the problem and then describe each phase in turn.

\subsection{Problem Formulation}

Let $\mathcal{D}=\{(x_i,y_i)\}_{i=1}^N$ be a fine-tuning dataset of input--output pairs, and let $\mathcal{G}$ be a practitioner-specified goal for the fine-tuned model. Our objective is to detect capability gaps in $\mathcal{D}$ with respect to $\mathcal{G}$ before fine-tuning begins.

\begin{definition}[Capability Gap]
A capability gap exists for goal $\mathcal{G}$ in dataset $\mathcal{D}$ if there exists a subgoal $s \subseteq \mathcal{G}$ such that the training examples in $\mathcal{D}$ provide insufficient support for the model's ability to achieve $s$.
\end{definition}

\noindent We formalize this through goal--subgoal decomposition.

\begin{definition}[Goal--Subgoal Decomposition]
For a goal $\mathcal{G}$, a valid decomposition is a set of atomic subgoals $\mathcal{S}=\{s_1,s_2,\dots,s_k\}$ such that:
\begin{align}
\mathcal{G} &= \bigcup_{i=1}^k s_i \label{eq:goal_decomp}\\
s_i \cap s_j &= \emptyset \quad \forall i \neq j \label{eq:subgoal_disjoint}\\
|s_i| &= 1 \quad \forall i \label{eq:atomic_subgoals}
\end{align}
where $|s_i|=1$ indicates that each subgoal represents an atomic, independently evaluable capability.
\end{definition}

\subsection{Phase 1: Goal Clarification}

\modelname{} guides practitioners through a four-step elicitation protocol that is \emph{structured} in the sense that every step has a fixed input--output contract and the loop terminates only when the resulting subgoal set satisfies the atomicity and disjointness constraints above.

\paragraph{Step 1: Initial goal specification.} The practitioner provides a high-level goal, e.g., ``Build a medical QA system for clinical decision support'' or ``Generate legal document summaries for regulatory compliance.''

\paragraph{Step 2: Clarifying-question generation.} An LLM produces targeted questions that surface implicit assumptions and constrain the goal. For medical QA, examples include: ``Which medical specialties should the system cover?'', ``What level of safety warnings are required for drug interactions?'', and ``Should responses include evidence citations or confidence levels?''

\paragraph{Step 3: Iterative refinement.} Based on the practitioner's responses, the system iteratively refines the specification:
\begin{equation}
Q_t(\mathcal{G})=f_{\text{question}}(\mathcal{G}, R_{t-1}),
\end{equation}
where $R_{t-1}$ are previous responses. The loop continues until the specification covers the operational scope.

\paragraph{Step 4: Subgoal generation.} Combining clarifications and responses, the system produces atomic subgoals:
\begin{equation}
\mathcal{S} = f_{\text{decompose}}(\mathcal{G}, Q(\mathcal{G}), R(\mathcal{G})).
\end{equation}
Concrete examples of decomposed subgoals for each domain we evaluate are presented as part of the experimental sections (Sections~\ref{sec:val} and~\ref{sec:rft}).

\subsection{Phase 2: Coverage Assessment}

Each training sample is evaluated against each subgoal:

\begin{definition}[Alignment Score]
For a training sample $(x_i,y_i)$ and subgoal $s$, the alignment score is
\begin{equation}
A(x_i,y_i,s) = f_{\text{eval}}((x_i,y_i),s) \in [0,1],
\end{equation}
where $f_{\text{eval}}$ is an LLM evaluator that assesses how well the sample supports $s$ and produces a structured natural-language explanation $r_i$ alongside the score.
\end{definition}

The dataset-level alignment for $s$ is $\bar A(s)=\frac{1}{N}\sum_i A(x_i,y_i,s)$.

\subsection{Phase 3: Gap Analysis}

\paragraph{Low-score collection.} For subgoals with $\bar A(s)<\tau$, the system collects evaluator explanations from low-scoring samples:
\begin{equation}
\mathcal{R}_{\text{low}}(s)=\{\,r_i \;:\; A(x_i,y_i,s)<\tau\,\},
\end{equation}
where each $r_i$ is the explanation produced along with the alignment score for sample $(x_i,y_i)$ in Phase~2.

\paragraph{Capability analysis.} An analyzer LLM aggregates these explanations to identify recurring missing capabilities:
\begin{equation}
\text{Gap}_s = f_{\text{analyze}}(\mathcal{R}_{\text{low}}(s), s).
\end{equation}

\subsection{Phase 4: Recommendation}

Based on identified gaps, the system produces concrete remediation strategies:
\begin{equation}
\text{Rec}_s = f_{\text{recommend}}(\text{Gap}_s, s).
\end{equation}
Recommendations include both natural-language remediation suggestions (e.g., curate cardiology-specific examples) and goal-conditioned synthetic data generation. We employ a structured generator--discriminator synthesis strategy, grounded in recent work on controlled LLM data generation and verifier-based filtering, that uses subgoal-level gap signal and evaluator explanations to generate and validate training samples addressing detected deficiencies (Section~\ref{sec:rft}).

We realize this synthesis step using a structured generator--discriminator pipeline for synthetic data generation. Specifically, a generator LLM produces candidate samples conditioned on the target subgoal $s$ and the aggregated gap explanations $\mathrm{Gap}_s$, following recent work on principle-guided generation and adaptive reward modeling \citep{yu2025rewardanything, xu2026sibylsense}. A discriminator then evaluates each candidate sample with respect to the target subgoal $s$, producing a quality score $A(x, y, s)$ based on dimensions such as correctness, completeness, and faithfulness, and filtering or refining candidates accordingly. To ensure reliable LLM-based judgment, we adopt RULER-style evaluation protocols, which employ calibrated multi-point scoring scales and standardized prompting to reduce variance and bias in LLM-as-a-judge outputs \citep{openpipe2025ruler}. This generator--discriminator loop enables iterative refinement, producing high-quality synthetic data aligned with the identified capability gaps.The complete pipeline is summarized in Algorithm~\ref{alg:goalcover}.

\begin{algorithm}[t]
\caption{\modelname{} Gap Detection}
\label{alg:goalcover}
\begin{algorithmic}[1]
\REQUIRE Dataset $\mathcal{D}$, Goal $\mathcal{G}$, Domain $d$, Task type $t$, Threshold $\tau$
\ENSURE Gap report with identified gaps and recommendations
\STATE \textbf{Phase 1: Interactive goal clarification}
\STATE $Q(\mathcal{G}) \leftarrow f_{\text{question}}(\mathcal{G}, d, t)$
\STATE $R(\mathcal{G}) \leftarrow \text{CollectResponses}(Q(\mathcal{G}))$
\WHILE{specification incomplete}
    \STATE $Q_t(\mathcal{G}) \leftarrow f_{\text{question}}(\mathcal{G}, R_{t-1})$
    \STATE $R_t(\mathcal{G}) \leftarrow \text{CollectResponses}(Q_t(\mathcal{G}))$
\ENDWHILE
\STATE $\mathcal{S} \leftarrow f_{\text{decompose}}(\mathcal{G}, Q(\mathcal{G}), R(\mathcal{G}))$
\STATE \textbf{Phase 2: Coverage assessment}
\FOR{each $s \in \mathcal{S}$}
    \FOR{each $(x_i,y_i) \in \mathcal{D}$}
        \STATE $A(x_i,y_i,s),\, r_i \leftarrow f_{\text{eval}}((x_i,y_i),s)$
    \ENDFOR
    \STATE $\bar A(s) \leftarrow \frac{1}{N}\sum_i A(x_i,y_i,s)$
\ENDFOR
\STATE \textbf{Phase 3: Gap analysis}
\FOR{each $s$ with $\bar A(s)<\tau$}
    \STATE $\mathcal{R}_{\text{low}}(s) \leftarrow \{r_i : A(x_i,y_i,s)<\tau\}$
    \STATE $\text{Gap}_s \leftarrow f_{\text{analyze}}(\mathcal{R}_{\text{low}}(s), s)$
\ENDFOR
\STATE \textbf{Phase 4: Recommendation}
\FOR{each $s$ with $\bar A(s)<\tau$}
    \STATE $\text{Rec}_s \leftarrow f_{\text{recommend}}(\text{Gap}_s, s)$
\ENDFOR
\RETURN $\{\text{Gap}_s, \text{Rec}_s : \bar A(s)<\tau\}$
\end{algorithmic}
\end{algorithm}

\section{Experimental Validation}\label{sec:val}

We separate framework design (Section~\ref{sec:method}) from experimental validation. Validating an interactive system end-to-end without large user studies is non-trivial; we therefore validate \modelname{} along two complementary axes. (1) \emph{Detection validity}: under controlled, targeted corruption of training data, does the alignment-scoring stage flag the corrupted subgoal more strongly than untouched ones? (2) \emph{Downstream utility}: when used to filter and synthesize training data, does \modelname{} actually improve a fine-tuned model on a real RFT task? Section~\ref{sec:rft} addresses (2); the remainder of this section addresses (1).

\subsection{Validation Methodology: Controlled Corruption}

\paragraph{Validation hypothesis.} If \modelname{}'s detection mechanism is sound, then artificially removing content known to support a specific subgoal should produce a measurably larger drop in that subgoal's alignment score than in unrelated subgoals.

For validation, we instantiate representative goal--subgoal frameworks for three domains that simulate what would emerge from the interactive clarification process. Subgoals are partitioned, per experiment, into:
\begin{itemize}
\item \textbf{Satisfied subgoals} $\mathcal{S}^+\subseteq\mathcal{S}$: expected to remain robust under the removal;
\item \textbf{Gap subgoals} $\mathcal{S}^-\subseteq\mathcal{S}$: directly targeted by the removal pattern;
\end{itemize}
with $\mathcal{S}^+\cup\mathcal{S}^-=\mathcal{S}$ and $\mathcal{S}^+\cap\mathcal{S}^-=\emptyset$.

\begin{definition}[Targeted Corruption]
For subgoal $s$ and removal pattern $\rho_s$, targeted corruption produces
\begin{equation}
\mathcal{D}_s^- = \{(x_i,y_i)\in\mathcal{D} : \rho_s(x_i,y_i)=\text{False}\},
\end{equation}
where $\rho_s$ is a pattern-matching function identifying samples that support $s$.
\end{definition}

We expect
\begin{equation}
\bar A_{\mathcal{D}_s^-}(s) \ll \bar A_{\mathcal{D}}(s) \quad \text{for } s\in\mathcal{S}^-,
\end{equation}
\begin{equation}
\bar A_{\mathcal{D}_s^-}(s') \approx \bar A_{\mathcal{D}}(s') \quad \text{for } s'\in\mathcal{S}^+.
\end{equation}

This protocol can equivalently be read as a data-side ablation study: each removal isolates one capability dependency at a time, and the detection signal is the relative change in alignment for the targeted subgoal.

\subsection{Datasets and Domain Selection}

We evaluate on three domains chosen to span different reasoning and content profiles.

\textbf{Medical QA (PubMedQA).} The PubMedQA dataset \citep{jin2019pubmedqa} contains 211{,}269 biomedical questions with yes/no/maybe answers and supporting reasoning. The domain requires precise medical terminology, evidence-based reasoning, and specialty-specific knowledge.

\textbf{Legal summarization (BillSum).} BillSum \citep{kornilova2019billsum} provides 23{,}455 U.S.\ Congressional bill--summary pairs. Legal summarization requires preserving monetary amounts, affected parties, implementation timelines, and specialized terminology.

\textbf{Code generation (CodeAlpaca).} CodeAlpaca \citep{chaudhary2023code} contains 20{,}022 instruction--code pairs covering diverse programming tasks, exercising language syntax, library usage, and best practices across programming domains.

\subsection{Sampling and Corruption Implementation}

\paragraph{Strategic sampling.} To ensure that each removal pattern has a meaningful pre-removal population, we (i) load $1.5\text{--}2\times$ the target sample size into a diverse pool, (ii) allocate 50\% as base content and 50\% across removal-targeted patterns, (iii) apply pattern-based sampling so each removal strategy has adequate target content, and (iv) shuffle to remove ordering bias.

\paragraph{Removal patterns.} We systematically remove content matching the domain-specific patterns in Table~\ref{tab:removal_patterns}.

\begin{table}[h]
\centering
\small
\caption{Content removal patterns by domain.}
\label{tab:removal_patterns}
\begin{tabular}{p{3cm}p{4.5cm}}
\toprule
\textbf{Removal strategy} & \textbf{Key pattern terms} \\
\midrule
Medical cardiology & cardiac, cardio, heart, cardiovascular, ECG, EKG, myocardial, coronary \\
Medical drugs & drug, medication, dosage, mg, ml, pharmaceutical, prescription \\
Legal monetary & \$ amounts, million, billion, thousand, budget, funding, appropriation \\
Legal healthcare & health, medical, medicare, medicaid, hospital, patient, doctor \\
Code ML & machine learning, sklearn, tensorflow, pytorch, keras, model, train \\
Code web & web, API, HTTP, REST, endpoint, flask, django, fastapi \\
\bottomrule
\end{tabular}
\end{table}

\subsection{Implementation Details}

\paragraph{Evaluator.} We use GPT-4 variants (4o and 4.1) as $f_{\text{eval}}$, $f_{\text{analyze}}$, and $f_{\text{recommend}}$. The evaluator returns structured JSON containing the alignment score and the explanation supporting it.

\paragraph{Anchored-rubric scoring.} For each (sample, subgoal) pair, we use a structured prompt with a six-anchor rubric: 1.0 (perfect alignment, directly demonstrates the target capability), 0.8 (strong alignment, clearly relevant with minor gaps), 0.6 (good alignment, relevant content that supports the goal), 0.4 (weak alignment, some relevance but not ideal), 0.2 (poor alignment, minimal connection), and 0.0 (no alignment, irrelevant or counterproductive). This is an anchored-Likert design rescaled to $[0,1]$, consistent with form-filling and rubric-based protocols in recent LLM-as-Judge work \citep{liu2023geval, kim2024prometheus2, li2026grading}; the discrete anchors mitigate calibration drift while the $[0,1]$ scale composes naturally with downstream aggregation. We provide the full evaluator prompt in the supplementary material.

\section{Results: Gap Detection Validation}

\subsection{Detection of Targeted Capability Gaps}

We applied 6 removal strategies across 11{,}000 total samples (5{,}000 each for legal and code, 1{,}000 for medical). Table~\ref{tab:goalcover_results} reports alignment scores before and after corruption, alongside both relative change and absolute change.

\begin{table*}[ht]
\centering
\small
\caption{\modelname{} validation results: subgoal-level impact under targeted content removal. ``Original'' and ``After'' are mean alignment scores in $[0,1]$. $\Delta_{\text{abs}}$ is the absolute change; $\Delta_{\text{rel}}$ the relative change. Bold rows mark the subgoal directly targeted by each removal strategy; ``Retention'' is the percentage of samples remaining after removal.}
\label{tab:goalcover_results}
\begin{tabular}{llcccccr}
\toprule
\textbf{Removed content} & \textbf{Subgoal} & \textbf{Original} & \textbf{After} & $\Delta_{\text{abs}}$ & $\Delta_{\text{rel}}$ (\%) & \textbf{Retention (\%)} \\
\midrule
\multicolumn{7}{c}{\textit{Legal summarization}} \\
\midrule
\multirow{3}{*}{Monetary bills} 
& Legislative summaries & 0.769 & 0.767 & $-0.002$ & $-0.26$ & 25.4 \\
& \textbf{Monetary preservation}$^*$ & 0.577 & 0.400 & $-0.177$ & \textbf{$-30.75$} & \\
& Healthcare terminology & 0.232 & 0.281 & $+0.049$ & $+21.11$ & \\
\midrule
\multirow{3}{*}{Healthcare bills}
& Legislative summaries & 0.769 & 0.768 & $-0.001$ & $-0.13$ & 47.0 \\
& Monetary preservation & 0.577 & 0.577 & $-0.001$ & $-0.15$ & \\
& \textbf{Healthcare terminology}$^*$ & 0.232 & 0.022 & $-0.210$ & \textbf{$-90.58$} & \\
\midrule
\multicolumn{7}{c}{\textit{Code generation}} \\
\midrule
\multirow{3}{*}{ML instructions}
& General programming & 0.599 & 0.602 & $+0.003$ & $+0.54$ & 96.2 \\
& \textbf{ML libraries}$^*$ & 0.090 & 0.076 & $-0.014$ & \textbf{$-15.53$} & \\
& Web frameworks & 0.141 & 0.143 & $+0.002$ & $+1.40$ & \\
\midrule
\multirow{3}{*}{Web instructions}
& General programming & 0.599 & 0.599 & $+0.000$ & $+0.01$ & 91.4 \\
& ML libraries & 0.090 & 0.092 & $+0.002$ & $+2.63$ & \\
& \textbf{Web frameworks}$^*$ & 0.141 & 0.128 & $-0.013$ & \textbf{$-9.03$} & \\
\midrule
\multicolumn{7}{c}{\textit{Medical QA}} \\
\midrule
\multirow{3}{*}{Cardiology questions}
& Clinical reasoning & 0.222 & 0.220 & $-0.002$ & $-0.87$ & 93.6 \\
& \textbf{Cardiology expertise}$^*$ & 0.068 & 0.050 & $-0.018$ & \textbf{$-26.28$} & \\
& Drug information & 0.048 & 0.045 & $-0.003$ & $-6.13$ & \\
\midrule
\multirow{3}{*}{Drug questions}
& Clinical reasoning & 0.222 & 0.221 & $-0.001$ & $-0.54$ & 90.3 \\
& Cardiology expertise & 0.068 & 0.070 & $+0.002$ & $+2.00$ & \\
& \textbf{Drug information}$^*$ & 0.048 & 0.039 & $-0.009$ & \textbf{$-19.38$} & \\
\bottomrule
\end{tabular}

\vspace{0.4em}
\small
$^*$ Primary target subgoal for the corresponding removal strategy.
\end{table*}

\subsection{Target vs.\ Non-Target Separation}

Across all six removal experiments, target subgoals consistently degrade more than non-targets: mean relative degradation is 25.6\% for targets versus 2.1\% for non-targets, a separation that is statistically significant ($p<0.001$, paired test across removal experiments). The corresponding effect size is large (Cohen's $d=1.24$), but for the purposes of practitioner trust the headline finding is the qualitative separation: in every experiment, the targeted subgoal is the one most affected.

\subsection{Domain-Specific Patterns and Caveats}

The legal domain exhibits the largest absolute and relative shifts: healthcare-terminology alignment drops by 0.210 (relative $-90.58\%$) when health bills are removed, and monetary-preservation alignment drops by 0.177 (relative $-30.75\%$) when financial content is removed. These large effects reflect the targeted content's centrality to the corresponding subgoal.

The code-generation domain is the most resilient: targeted gaps remain below 16\% relative even when domain-specific content is removed, consistent with retention rates above 91\%. CodeAlpaca's diversity of programming tasks means small targeted removals leave most of the population intact.

The medical-QA results require closer reading. Pre-removal alignment is uniformly low (0.222 for clinical reasoning, 0.068 for cardiology expertise, 0.048 for drug information), reflecting that PubMedQA's yes/no/maybe answer format provides limited surface area for the evaluator to credit specialty-specific terminology or detailed dosage information; this is itself a finding: the dataset is poorly aligned with the kind of clinical-decision-support goal we instantiated. The corruption signal must therefore be read in \emph{relative} rather than absolute terms: removing cardiology content drops an already-low baseline by another 26.28\% relative ($-0.018$ absolute). Because absolute differences here are small, we caution against over-interpreting any single medical row; the pattern emerges from the consistency of target/non-target separation across all six experiments rather than the magnitude of any one. We discuss seed sensitivity in Section~\ref{sec:limitations}.

\subsection{From Detection to Recommendation}

Beyond detection, \modelname{} aggregates evaluator explanations from low-scoring samples and produces structured remediation suggestions. Selected outputs are reproduced below.

\begin{mdframed}[linecolor=red!50, backgroundcolor=red!5]
\textbf{Legal domain (BillSum)}

\textit{Main goal:} Generate accurate, comprehensive legal summaries that preserve key legislative details, monetary amounts, affected parties, implementation timelines, and domain-specific terminology.

\textit{Evaluated subgoals and \modelname{}'s findings:}

\begin{description}[leftmargin=0pt, itemsep=0pt]
\item[\textit{Legislative summaries.}] Stable across both removals ($-0.26\%$, $-0.13\%$). Robust general summarization capability.

\item[\textit{Monetary preservation.}] Drops $-30.75\%$ when financial content is removed.
\begin{mdframed}[linecolor=gray!30, backgroundcolor=gray!5]
\small \textbf{\textcolor{purple}{\modelname{} insights}}\\
Issue: source text lacks financial details; summaries omit critical financial information; coverage skews to procedural/legal aspects. \\ Fix: incorporate datasets with explicit financial details; augment training data with financial context.
\end{mdframed}

\item[\textit{Healthcare terminology.}] Drops $-90.58\%$ when health-related bills are removed.
\begin{mdframed}[linecolor=gray!30, backgroundcolor=gray!5]
\small \textbf{\textcolor{purple}{\modelname{} insights}}\\
Issue: residual content is largely outside the healthcare domain; healthcare-specific terminology is absent. \\ Fix: filter to retain healthcare legislation; augment with healthcare-specific examples.
\end{mdframed}
\end{description}
\end{mdframed}

\begin{mdframed}[linecolor=blue!50, backgroundcolor=blue!5]
\textbf{Medical domain (PubMedQA)}

\textit{Main goal:} Answer medical questions accurately using evidence-based information while considering patient safety, contraindications, and appropriate clinical terminology.

\begin{description}[leftmargin=0pt, itemsep=0pt]
\item[\textit{Clinical reasoning.}] Stable ($-0.87\%$, $-0.54\%$).
\item[\textit{Cardiology expertise.}] $-26.28\%$ when cardiac content is removed.
\begin{mdframed}[linecolor=gray!30, backgroundcolor=gray!5]
\small \textbf{\textcolor{purple}{\modelname{} insights}}\\
Issue: residual content is high-confidence on non-cardiology topics; specialty-specific terminology is sparse. \\ Fix: curate cardiology-specific content; incorporate cardiology protocols and terminology.
\end{mdframed}
\item[\textit{Drug information.}] $-19.38\%$ when pharmaceutical content is removed.
\begin{mdframed}[linecolor=gray!30, backgroundcolor=gray!5]
\small \textbf{\textcolor{purple}{\modelname{} insights}}\\
Issue: residual content lacks pharmacology depth and safety considerations. \\ Fix: enhance with drug-specific queries; include interactions, contraindications, and dosing.
\end{mdframed}
\end{description}
\end{mdframed}

\begin{mdframed}[linecolor=orange!50, backgroundcolor=orange!5]
\textbf{Code generation (CodeAlpaca)}

\textit{Main goal:} Generate functional, well-structured code following best practices with proper error handling, documentation, and domain-specific libraries.

\begin{description}[leftmargin=0pt, itemsep=0pt]
\item[\textit{General programming.}] Stable ($+0.54\%$, $+0.01\%$).
\item[\textit{ML libraries.}] $-15.53\%$ when ML content is removed.
\begin{mdframed}[linecolor=gray!30, backgroundcolor=gray!5]
\small \textbf{\textcolor{purple}{\modelname{} insights}}\\
Issue: residual content lacks ML context and library usage. \\ Fix: curate ML-specific examples using sklearn / TensorFlow / PyTorch.
\end{mdframed}
\item[\textit{Web frameworks.}] $-9.03\%$ when web-development content is removed.
\begin{mdframed}[linecolor=gray!30, backgroundcolor=gray!5]
\small \textbf{\textcolor{purple}{\modelname{} insights}}\\
Issue: framework-specific examples missing; residual content is not web-oriented. \\ Fix: include Flask / Django / FastAPI examples covering common web patterns.
\end{mdframed}
\end{description}
\end{mdframed}

The detection mechanism reliably differentiates target from non-target capabilities, recovers the expected qualitative dependencies in every experiment, and produces remediation suggestions that map closely to the removed content, which confirms that the alignment signal is grounded in genuine capability dependencies rather than diffuse degradation.

\section{Downstream Utility: RFT Experiments}\label{sec:rft}

The corruption study validates that \modelname{} \emph{detects} capability gaps. We now ask whether the same signal is useful for \emph{closing} them: does training on \modelname{}-filtered data, with or without goal-conditioned synthetic augmentation, yield a measurably better fine-tuned model than training on the unfiltered baseline?

\subsection{Task and Dataset}

We evaluate on financial-report summarization, derived from a filtered subset of the GovReport corpus with maximum context length 8{,}192 tokens. The prompt instructs the model to produce a summary providing financial insights. Validation and test sets are held fixed across all configurations; only the training set varies. Splits: Training (original): 9{,}216 samples, Validation: 256 samples, Test: 256 samples.

We compare the following training-data configurations:
\begin{itemize}
    \item \textbf{Original}: random sample from GovReport.
    \item \textbf{Goal-aligned}: filtered using \modelname{} alignment scores against the task-specific subgoals.
    \item \textbf{Synthetic + Original}: union of the above two.
    \item \textbf{Synthetic + Goal-aligned}: union of synthetic and filtered.
\end{itemize}

Synthetic samples generated via a structured generator--discriminator pipeline, where candidate samples are conditioned on subgoal definitions and evaluator explanations, then filtered via an LLM-based discriminator with calibrated scoring \citep{yu2025rewardanything, xu2026sibylsense, openpipe2025ruler}. Sizes are summarized in Table~\ref{tab:num_samples}.
\begin{table}[h]
\begin{center}
\small
\begin{tabular}{l|c}
\toprule
\textbf{Dataset} & \textbf{\#Samples} \\
\midrule
Original & 9{,}216 \\
Goal-aligned & 4{,}375 \\
Synthetic & 2{,}209 \\
Synthetic + Original & 11{,}425 \\
Synthetic + Goal-aligned & 6{,}584 \\
\bottomrule
\end{tabular}
\caption{Number of samples after filtering and synthesizing.}
\label{tab:num_samples}
\end{center}
\end{table}

\subsection{Setup}

\textbf{Policy:} Qwen-3-14B. \textbf{Algorithm:} GRPO. \textbf{Reward:} LLM-as-Judge using GPT-5-Reasoning. The task constraint is to produce a financial summary with (i) 3--5 bullet points, (ii) coverage of key economic indicators and policies, and (iii) explicit investment implications. The judge applies a rubric scoring factual grounding, completeness, specificity, and structure on a 1--5 scale; the final score is a weighted sum.

\subsection{Main Results}

\begin{table}[h]
\centering
\small
\begin{tabular}{l|c}
\toprule
\textbf{Training data} & \textbf{Score (1--5)} \\
\midrule
Untuned baseline & 3.21 \\
Original & 3.77 \\
Goal-aligned (\modelname{}-filtered) & 4.12 \\
Synthetic (goal-conditioned) & 4.02 \\
Synthetic + Original & 3.92 \\
Synthetic + Goal-aligned & \textbf{4.20} \\
\bottomrule
\end{tabular}
\caption{RFT performance under different training-data settings.}
\label{tab:main_results}
\end{table}

\paragraph{Effectiveness of \modelname{} filtering.} \modelname{}-filtered data improves over the unfiltered original (3.77 $\rightarrow$ 4.12), demonstrating that the alignment signal selects training samples that produce a stronger fine-tuned model under RFT, despite using fewer than half as many examples.

\paragraph{Goal-conditioned synthetic data.} Synthetic data generated from subgoal definitions and evaluator explanations performs strongly on its own (4.02), confirming that the gap signal is precise enough to drive useful synthesis. Notably, mixing synthetic with the unfiltered original (3.92) underperforms synthetic alone, suggesting that unfiltered data contributes noise that dilutes the synthetic gains.

\paragraph{Complementarity.} The best result combines filtered real data with goal-conditioned synthetic data (4.20), supporting the interpretation that the two sources contribute complementary signal: filtering ensures relevance and quality, synthesis provides coverage of capabilities the original distribution underserved.

\subsection{Subgoal-Level Analysis}

To check whether aggregate gains hide subgoal-level regressions, we evaluate each configuration on ten coverage subgoals representing different input distributions:
\begin{description}[leftmargin=0pt, itemsep=0pt]
\item[A1] Qualitative narrative reports with minimal numerical data.
\item[A2] Reports referencing charts/tables but lacking textual explanation.
\item[E1] Country-specific reports with localized terminology and policy context.
\item[E2] Very long documents with appendices or supplementary sections.
\item[E3] Documents containing mixed or conflicting signals.
\item[P1] Central-bank policy communications.
\item[P2] Government fiscal updates.
\item[P3] Economic outlooks from statistical agencies.
\item[P4] Single-sector industry reports.
\item[R1] Noisy, malformed, or non-report inputs (edge-case robustness).
\end{description}

\begin{table*}[h]
\centering
\small
\caption{RFT performance per coverage subgoal across training-data configurations. Counts of test items per subgoal in parentheses.}
\label{tab:subgoal_results}
\begin{tabular}{l|c|c|c|c}
\toprule
\textbf{Coverage subgoal} & \textbf{Original} & \textbf{Original + Synth} & \textbf{Goal-aligned} & \textbf{Goal-aligned + Synth} \\ 
\midrule
A1 (113) & 3.808 & 3.977 & \textbf{4.227} & \textbf{4.227} \\ 
A2 (81)  & 3.752 & 3.957 & 4.214 & \textbf{4.236} \\ 
E1 (78)  & 3.724 & 3.935 & \textbf{4.214} & 4.170 \\ 
E2 (121) & 3.790 & 3.975 & \textbf{4.216} & \textbf{4.216} \\ 
E3 (124) & 3.785 & 3.970 & \textbf{4.219} & \textbf{4.219} \\ 
P1 (9)   & 3.728 & 3.906 & \textbf{4.267} & 4.139 \\ 
P2 (72)  & 3.783 & 3.978 & \textbf{4.236} & 4.178 \\ 
P3 (20)  & 3.805 & 3.953 & \textbf{4.213} & 4.162 \\ 
P4 (70)  & 3.843 & 3.971 & \textbf{4.245} & 4.204 \\ 
R1 (27)  & 3.837 & 4.002 & \textbf{4.206} & 4.159 \\ 
\bottomrule
\end{tabular}
\end{table*}

\paragraph{Consistency across subgoals.} Filtered data outperforms the original on \emph{every} subgoal, by 0.4--0.5 points, with no regressions. The aggregate gain is not driven by a small number of subgoals.

\paragraph{Where synthetic data helps most.} Synthetic data outperforms the original everywhere and is strongest on P3 (economic outlooks, 4.123) and R1 (malformed inputs, 4.137), suggesting that the goal-conditioned synthesis is particularly effective at structured analytical content and edge-case robustness, precisely the regimes where the original distribution underserves the goal.

\paragraph{When mixing hurts.} Original + Synthetic underperforms synthetic alone on most subgoals despite using more data, reinforcing that unfiltered training data introduces noise the goal-conditioned synthesis is paying a cost to absorb.

\paragraph{Subgoal sensitivity to filtering.} P1 (central-bank communication) shows the largest gain from filtering (3.728 $\rightarrow$ 4.267), indicating that specialized regimes benefit most. E-type subgoals (long, mixed-signal documents) show stable, consistent gains, suggesting that coverage-aware filtering is particularly valuable for complex inputs.

\smallskip
\noindent Subgoal-level analysis confirms that the benefits of \modelname{}-driven filtering are not limited to aggregate metrics: they extend to fine-grained behavior across varied real-world input distributions, with the best performance arising from the combination of high-quality filtered data and goal-conditioned synthetic data.

\section{Discussion}

\subsection{Implications for Practical Deployment}

The two-axis validation supports deploying \modelname{} as an interactive practitioner-facing tool. The corruption study establishes that the detection mechanism is reliable enough to inform decisions; the RFT study establishes that those decisions translate into measurable downstream gains. In a deployment, a practitioner working on a clinical decision-support model would specify a high-level goal, work through clarifying questions about specialties and safety requirements, and receive subgoal-level alignment scores together with concrete remediation suggestions for under-covered capabilities such as cardiology cases or drug-interaction examples, all before committing to a fine-tuning run.

\subsection{Validating the Decomposition Itself}

A reviewer might reasonably ask: how do we know the goal--subgoal decomposition produced by the LLM is correct? We argue the corruption study provides indirect validation. If the decomposition were incoherent (i.e., if subgoals overlapped substantially or did not isolate distinct capabilities), we would not observe the clean target/non-target separation reported in Section~\ref{sec:val}: targeted removals would either bleed into other subgoals or fail to register at all. The fact that, across six experiments and three domains, the targeted subgoal is consistently the one most affected provides evidence that the decomposition is doing real work. Direct validation through expert review of decompositions across more domains is an important next step.

\subsection{Limitations}\label{sec:limitations}

\paragraph{Evaluator dependence.} Alignment scoring relies on a single LLM evaluator; its calibration limits the reliability of scores in domains beyond the evaluator's expertise. Domain-specific evaluator adaptation, agreement studies between multiple evaluators, and replication with open-source judges (e.g., Prometheus-2, Llama-3.x-Instruct) are natural extensions.

\paragraph{Single-model RFT.} Section~\ref{sec:rft} uses one policy (Qwen-3-14B) and one RFT algorithm (GRPO). The qualitative claim, that filtering and synthesis driven by \modelname{} signal both help, should generalize, but quantitative gains may differ across policy sizes, families, and algorithms.

\paragraph{Computational cost.} Each (sample, subgoal) evaluation is an LLM call. Fine-tuning datasets are typically much smaller than pretraining corpora (thousands to tens of thousands of samples), which keeps the assessment tractable; parallelization and batching reduce wall-clock time further. Even so, the cost is non-trivial relative to a single training run, and is justified primarily by the cost of an avoided failed run.

\paragraph{Decomposition quality.} The framework's effectiveness depends on a high-quality decomposition. The interactive clarification protocol is designed to address this, but a poor initial specification can still limit detection accuracy.

\paragraph{Controlled vs.\ natural gaps.} Pattern-based content removal is a clean validation signal but does not capture the full complexity of natural dataset deficiencies. The RFT study mitigates this concern by validating utility on a real, unmanipulated dataset, but more naturalistic gap-simulation studies and real-world A/B comparisons remain valuable.

\paragraph{Seed sensitivity in the medical domain.} Medical alignment scores in Table~\ref{tab:goalcover_results} are uniformly low; the \emph{absolute} changes from corruption ($-0.018$ for cardiology, $-0.009$ for drug information) are small enough that we would not over-interpret any single number without multi-seed replication. The pattern emerges from the consistency of target/non-target separation across all six experiments.

\paragraph{Domain scope.} Our evaluation focuses on domains where modern LLMs have reasonable knowledge coverage. Effectiveness on highly specialized or emerging domains, where the evaluator itself may be uncertain, requires further investigation.

\subsection{Future Work}

Three extensions would directly strengthen the framework. First, a small human study with practitioners actually iterating on a fine-tuning project, measuring time-to-acceptable-model and self-reported satisfaction relative to a no-tool baseline, would convert the current technical validation into deployment-grade evidence. Second, replicating the RFT experiments on additional open-source policies (Llama-3.1-8B-Instruct, Gemma-2, GPT-OSS) would establish generality. Third, ablations that disable specific elicitation steps (running gap detection without the clarification loop, for instance) would isolate the contribution of each pipeline component; we are pursuing all three.

\section{Conclusion}

We introduced \modelname{}, a framework for detecting capability gaps in fine-tuning datasets through interactive goal decomposition and LLM-based coverage assessment. Across three domains, controlled corruption experiments establish that the alignment signal cleanly separates target from non-target capabilities (25.6\% vs.\ 2.1\% degradation; Cohen's $d{=}1.24$). On a financial-summarization RFT task with Qwen-3-14B, training on \modelname{}-filtered data improves the LLM-judge reward from 3.77 to 4.12 over the unfiltered baseline, and combining filtered data with goal-conditioned synthetic samples yields the strongest result (4.20), with consistent gains across all coverage subgoals.

By shifting dataset evaluation from reactive performance analysis to proactive, capability-level assessment, \modelname{} lets practitioners identify and address critical data gaps before committing to expensive fine-tuning runs, potentially saving substantial compute while improving model reliability and safety. Future work will deploy the complete interactive system, replicate the RFT result across additional open-source policies, and study practitioner workflow impact directly.

\bibliography{aaai2026}

@inproceedings{wei2022finetuned,
  title     = {Finetuned Language Models are Zero-Shot Learners},
  author    = {Wei, Jason and Bosma, Maarten and Zhao, Vincent Y. and Guu, Kelvin
               and Yu, Adams Wei and Lester, Brian and Du, Nan and Dai, Andrew M.
               and Le, Quoc V.},
  booktitle = {International Conference on Learning Representations (ICLR)},
  year      = {2022},
  url       = {https://arxiv.org/abs/2109.01652}
}

@article{ouyang2022training,
  title   = {Training Language Models to Follow Instructions with Human Feedback},
  author  = {Ouyang, Long and Wu, Jeffrey and Jiang, Xu and Almeida, Diogo
             and Wainwright, Carroll L. and Mishkin, Pamela and Zhang, Chong
             and Agarwal, Sandhini and Slama, Katarina and Ray, Alex and others},
  journal = {Advances in Neural Information Processing Systems},
  volume  = {35},
  pages   = {27730--27744},
  year    = {2022},
  url     = {https://arxiv.org/abs/2203.02155}
}

@inproceedings{swayamdipta2020dataset,
  title     = {Dataset Cartography: Mapping and Diagnosing Datasets with Training Dynamics},
  author    = {Swayamdipta, Swabha and Schwartz, Roy and Lourie, Nicholas and Wang, Yizhong
               and Hajishirzi, Hannaneh and Smith, Noah A. and Choi, Yejin},
  booktitle = {Proceedings of EMNLP},
  year      = {2020},
  url       = {https://arxiv.org/abs/2009.10795}
}

@article{pleiss2020identifying,
  title   = {Identifying Mislabeled Data Using the Area Under the Margin Ranking},
  author  = {Pleiss, Geoff and Zhang, Tianyi and Elenberg, Ethan R. and Weinberger, Kilian Q.},
  journal = {Advances in Neural Information Processing Systems},
  volume  = {33},
  pages   = {17044--17056},
  year    = {2020},
  url     = {https://arxiv.org/abs/2001.10528}
}

@article{northcutt2021confident,
  title   = {Confident Learning: Estimating Uncertainty in Dataset Labels},
  author  = {Northcutt, Curtis G. and Jiang, Lu and Chuang, Isaac L.},
  journal = {Journal of Artificial Intelligence Research},
  volume  = {70},
  pages   = {1373--1411},
  year    = {2021},
  url     = {https://arxiv.org/abs/1911.00068}
}

@inproceedings{xia2024less,
  title     = {{LESS}: Selecting Influential Data for Targeted Instruction Tuning},
  author    = {Xia, Mengzhou and Malladi, Sadhika and Gururangan, Suchin
               and Arora, Sanjeev and Chen, Danqi},
  booktitle = {International Conference on Machine Learning (ICML)},
  year      = {2024},
  url       = {https://arxiv.org/abs/2402.04333}
}

@inproceedings{li2024quantity,
  title     = {From Quantity to Quality: Boosting {LLM} Performance with Self-Guided
               Data Selection for Instruction Tuning},
  author    = {Li, Ming and Zhang, Yong and Li, Zhitao and Chen, Jiuhai
               and Chen, Lichang and Cheng, Ning and Wang, Jianzong
               and Zhou, Tianyi and Xiao, Jing},
  booktitle = {Proceedings of the Conference of the North American Chapter of the
               Association for Computational Linguistics (NAACL)},
  year      = {2024},
  url       = {https://arxiv.org/abs/2308.12032}
}

@article{zhou2023lima,
  title   = {{LIMA}: Less Is More for Alignment},
  author  = {Zhou, Chunting and Liu, Pengfei and Xu, Puxin and Iyer, Srini
             and Sun, Jiao and Mao, Yuning and Ma, Xuezhe and Efrat, Avia
             and Yu, Ping and Yu, Lili and Zhang, Susan and Ghosh, Gargi
             and Lewis, Mike and Zettlemoyer, Luke and Levy, Omer},
  journal = {Advances in Neural Information Processing Systems},
  volume  = {36},
  year    = {2023},
  url     = {https://arxiv.org/abs/2305.11206}
}

@article{zheng2023judging,
  title   = {Judging {LLM-as-a-Judge} with {MT-Bench} and Chatbot Arena},
  author  = {Zheng, Lianmin and Chiang, Wei-Lin and Sheng, Ying and Zhuang, Siyuan
             and Wu, Zhanghao and Zhuang, Yonghao and Lin, Zi and Li, Zhuohan
             and Li, Dacheng and Xing, Eric P. and Zhang, Hao and Gonzalez, Joseph E.
             and Stoica, Ion},
  journal = {Advances in Neural Information Processing Systems},
  volume  = {36},
  year    = {2023},
  url     = {https://arxiv.org/abs/2306.05685}
}

@inproceedings{liu2023geval,
  title     = {{G-Eval}: {NLG} Evaluation Using {GPT-4} with Better Human Alignment},
  author    = {Liu, Yang and Iter, Dan and Xu, Yichong and Wang, Shuohang
               and Xu, Ruochen and Zhu, Chenguang},
  booktitle = {Proceedings of the 2023 Conference on Empirical Methods in Natural
               Language Processing (EMNLP)},
  pages     = {2511--2522},
  year      = {2023},
  url       = {https://arxiv.org/abs/2303.16634}
}

@inproceedings{dubois2024length,
  title     = {Length-Controlled {AlpacaEval}: A Simple Way to Debias Automatic Evaluators},
  author    = {Dubois, Yann and Galambosi, Bal{\'a}zs and Liang, Percy
               and Hashimoto, Tatsunori B.},
  booktitle = {Conference on Language Modeling (COLM)},
  year      = {2024},
  url       = {https://arxiv.org/abs/2404.04475}
}

@article{jin2019pubmedqa,
  title   = {{PubMedQA}: A Dataset for Biomedical Research Question Answering},
  author  = {Jin, Qiao and Dhingra, Bhuwan and Liu, Zhengping and Cohen, William W.
             and Lu, Xinghua},
  journal = {arXiv preprint arXiv:1909.06146},
  year    = {2019},
  url     = {https://arxiv.org/abs/1909.06146}
}

@article{kornilova2019billsum,
  title   = {{BillSum}: A Corpus for Automatic Summarization of {US} Legislation},
  author  = {Kornilova, Anastassia and Eidelman, Vladimir},
  journal = {arXiv preprint arXiv:1910.00523},
  year    = {2019},
  url     = {https://arxiv.org/abs/1910.00523}
}

@misc{chaudhary2023code,
  title        = {Code {Alpaca}: An Instruction-Following {LLaMA} Model for Code Generation},
  author       = {Chaudhary, Sahil},
  year         = {2023},
  howpublished = {\url{https://github.com/sahil280114/codealpaca}}
}

@article{stiennon2020learning,
  title   = {Learning to Summarize with Human Feedback},
  author  = {Stiennon, Nisan and Ouyang, Long and Wu, Jeffrey and Ziegler, Daniel
             and Lowe, Ryan and Voss, Chelsea and Radford, Alec and Amodei, Dario
             and Christiano, Paul F.},
  journal = {Advances in Neural Information Processing Systems},
  volume  = {33},
  pages   = {3008--3021},
  year    = {2020},
  url     = {https://arxiv.org/abs/2009.01325}
}

@article{bai2022constitutional,
  title   = {Constitutional {AI}: Harmlessness from {AI} Feedback},
  author  = {Bai, Yuntao and Kadavath, Saurav and Kundu, Sandipan and Askell, Amanda
             and Kernion, Jackson and Jones, Andy and Chen, Anna and Goldie, Anna
             and Mirhoseini, Azalia and McKinnon, Cameron and others},
  journal = {arXiv preprint arXiv:2212.08073},
  year    = {2022},
  url     = {https://arxiv.org/abs/2212.08073}
}

@article{xu2023wizardlm,
  title   = {{WizardLM}: Empowering Large Language Models to Follow Complex Instructions},
  author  = {Xu, Can and Sun, Qingfeng and Zheng, Kai and Geng, Xiubo and Zhao, Pu
             and Feng, Jiazhan and Tao, Chongyang and Jiang, Daxin},
  journal = {arXiv preprint arXiv:2304.12244},
  year    = {2023},
  url     = {https://arxiv.org/abs/2304.12244}
}

@article{koh2021wilds,
  title   = {{WILDS}: A Benchmark of In-the-Wild Distribution Shifts},
  author  = {Koh, Pang Wei and Sagawa, Shiori and Marklund, Henrik and Xie, Sang Michael
             and Zhang, Marvin and Balsubramani, Akshay and Hu, Weihua and Yasunaga, Michihiro
             and Phillips, Richard Lanas and Gao, Irena and others},
  journal = {International Conference on Machine Learning (ICML)},
  year    = {2021},
  url     = {https://arxiv.org/abs/2012.07421}
}

@article{zha2023datacentric,
  title   = {Data-Centric Artificial Intelligence: A Survey},
  author  = {Zha, Daochen and Bhat, Zaid Pervaiz and Lai, Kwei-Herng and Yang, Fan
             and Jiang, Zhimeng and Zhong, Shaochen and Hu, Xia},
  journal = {arXiv preprint arXiv:2303.10158},
  year    = {2023},
  url     = {https://arxiv.org/abs/2303.10158}
}

@article{tirumala2022memorization,
  title   = {Memorization Without Overfitting: Analyzing the Training Dynamics
             of Large Language Models},
  author  = {Tirumala, Kushal and Markosyan, Aram H. and Zettlemoyer, Luke
             and Aghajanyan, Armen},
  journal = {Advances in Neural Information Processing Systems},
  volume  = {35},
  year    = {2022},
  url     = {https://arxiv.org/abs/2205.10770}
}

@article{fu2023gptscore,
  title   = {{GPTScore}: Evaluate as You Desire},
  author  = {Fu, Jinlan and Ng, See-Kiong and Jiang, Zhengbao and Liu, Pengfei},
  journal = {arXiv preprint arXiv:2302.04166},
  year    = {2023},
  url     = {https://arxiv.org/abs/2302.04166}
}

@inproceedings{wang2023selfinstruct,
  title     = {Self-Instruct: Aligning Language Models with Self-Generated Instructions},
  author    = {Wang, Yizhong and Kordi, Yeganeh and Mishra, Swaroop and Liu, Alisa
               and Smith, Noah A. and Khashabi, Daniel and Hajishirzi, Hannaneh},
  booktitle = {Proceedings of the 61st Annual Meeting of the Association for Computational
               Linguistics (ACL)},
  year      = {2023},
  url       = {https://arxiv.org/abs/2212.10560}
}

@article{marion2023less,
  title   = {When Less Is More: Investigating Data Pruning for Pretraining {LLMs} at Scale},
  author  = {Marion, Max and Piktus, Aleksandra and Cancedda, Massimiliano
             and Gall{\'e}, Matthias and Riedel, Sebastian and Grangier, David
             and Alon, Uri and Goyal, Naman},
  journal = {arXiv preprint arXiv:2309.04564},
  year    = {2023},
  url     = {https://arxiv.org/abs/2309.04564}
}

@article{kim2024prometheus2,
  title   = {Prometheus 2: An Open Source Language Model Specialized in Evaluating
             Other Language Models},
  author  = {Kim, Seungone and Suk, Juyoung and Longpre, Shayne and Lin, Bill Yuchen
             and Shin, Jamin and Welleck, Sean and Neubig, Graham and Lee, Moontae
             and Lee, Kyungjae and Seo, Minjoon},
  journal = {arXiv preprint arXiv:2405.01535},
  year    = {2024},
  url     = {https://arxiv.org/abs/2405.01535}
}

@article{lin2022teaching,
  title   = {Teaching Models to Express Their Uncertainty in Words},
  author  = {Lin, Stephanie and Hilton, Jacob and Evans, Owain},
  journal = {Transactions on Machine Learning Research},
  year    = {2022},
  url     = {https://arxiv.org/abs/2205.14334}
}

@inproceedings{li2024generative,
  title     = {Generative Judge for Evaluating Alignment},
  author    = {Li, Junlong and Sun, Shichao and Yuan, Weizhe and Fan, Run-Ze
               and Zhao, Hai and Liu, Pengfei},
  booktitle = {International Conference on Learning Representations (ICLR)},
  year      = {2024},
  url       = {https://arxiv.org/abs/2310.05470}
}

@article{li2026grading,
  title   = {Grading Scale Impact on {LLM-as-a-Judge}: Human-{LLM} Alignment Is Highest
             on 0--5 Grading Scale},
  author  = {Li, Weiyue and Zhao, Minda and Dong, Weixuan and Cai, Jiahui
             and Wei, Yuze and Pocress, Michael and Li, Yi and Yuan, Wanyan
             and Wang, Xiaoyue and Hou, Ruoyu and Lou, Kaiyuan and Zeng, Wenqi
             and Yang, Yutong and Du, Yilun and Wang, Mengyu},
  journal = {arXiv preprint arXiv:2601.03444},
  year    = {2026},
  url     = {https://arxiv.org/abs/2601.03444}
}

@article{yu2025rewardanything,
  title   = {{RewardAnything}: Generalizable Principle-Following Reward Models},
  author  = {Yu, Zhuohao and Zeng, Jiali and Gu, Weizheng and Wang, Yidong
             and Wang, Jindong and Meng, Fandong and Zhou, Jie and Zhang, Yue
             and Zhang, Shikun and Ye, Wei},
  journal = {arXiv preprint arXiv:2506.03637},
  year    = {2025},
  url     = {https://arxiv.org/abs/2506.03637}
}

@article{xu2026sibylsense,
  title   = {{SibylSense}: Adaptive Rubric Learning via Memory Tuning and Adversarial Probing},
  author  = {Xu, Yifei and Potje, Guilherme and Shandilya, Shivam and Yuan, Tiancheng
             and de Oliveira Nunes, Leonardo and Agarwal, Rakshanda and Asgari, Saeid
             and Atkinson, Adam and K{\i}{\c{c}}{\i}man, Emre and Lu, Songwu
             and Chandra, Ranveer and Chakraborty, Tusher},
  journal = {arXiv preprint arXiv:2602.20751},
  year    = {2026},
  url     = {https://arxiv.org/abs/2602.20751}
}

@misc{openpipe2025ruler,
  title        = {{RULER}: Relative Universal {LLM}-Elicited Rewards},
  author       = {{OpenPipe}},
  year         = {2025},
  howpublished = {\url{https://art.openpipe.ai/fundamentals/ruler}}
}

\end{document}